\documentclass[10pt,twocolumn,letterpaper]{article}

\usepackage[pagenumbers]{cvpr} %

\usepackage{graphicx}
\usepackage{amsmath}
\usepackage{amssymb}
\usepackage{booktabs}
\usepackage{stackengine}
\usepackage{amsthm}
\newtheorem{constraint}{Constraint}
\newtheorem{definition}{Definition}
\newtheorem{proposition}{Proposition}

\usepackage[pagebackref,breaklinks,colorlinks]{hyperref}

\usepackage[capitalize]{cleveref}
\crefname{section}{Sec.}{Secs.}
\Crefname{section}{Section}{Sections}
\Crefname{table}{Table}{Tables}
\crefname{table}{Tab.}{Tabs.}

\begin{document}

\title{Stable Long-Term Recurrent Video Super-Resolution}

\author{Benjamin Naoto Chiche $^{1, 2}$\\
{\tt\small benjamin.naoto.chiche@outlook.com}
\and
Arnaud Woiselle $^1$\\
{\tt\small arnaud.woiselle@safrangroup.com}
\and
Joana Frontera-Pons $^{2, 3}$\\
{\tt\small joana.frontera-pons@cea.fr}
\and
Jean-Luc Starck $^2$\\
{\tt\small https://orcid.org/0000-0003-2177-7794}
\and
\small $^1$ Safran Electronics \& Defense\\
\small F-91344 Massy, France\\
\small $^2$ AIM, CEA, CNRS, Université Paris-Saclay, Université de Paris \\
\small F-91191 Gif-sur-Yvette, France\\
\small $^3$ DR2I, Institut Polytechnique des Sciences Avancées\\
\small F-94200 Ivry-sur-Seine, France \\
\and
}
\maketitle

\begin{abstract}

Recurrent models have gained popularity in deep learning (DL) based video super-resolution (VSR), due to their increased computational efficiency, temporal receptive field and temporal consistency compared to sliding-window based models. However, when inferring on long video sequences presenting low motion (\ie~in which some parts of the scene barely move), recurrent models diverge through recurrent processing, generating high frequency artifacts. To the best of our knowledge, no study about VSR pointed out this instability problem, which can be critical for some real-world applications. Video surveillance is a typical example where such artifacts would occur, as both the camera and the scene stay static for a long time. 

In this work, we expose instabilities of existing recurrent VSR networks on long sequences with low motion. We demonstrate it on a new long sequence dataset Quasi-Static Video Set, that we have created. Finally, we introduce a new framework of recurrent VSR networks that is both stable and competitive, based on Lipschitz stability theory. We propose a new recurrent VSR network, coined Middle Recurrent Video Super-Resolution (MRVSR), based on this framework. We empirically show its competitive performance on long sequences with low motion.

\end{abstract}

\begin{figure}[!htbp]

\centering

\begin{subfigure}{0.475\columnwidth} 
\stackinset{r}{-.\textwidth}{t}{-.\textwidth}
   {
   \setlength{\fboxsep}{0pt}%
 \setlength{\fboxrule}{1pt}%
   \fbox{\includegraphics[width=0.4\textwidth]{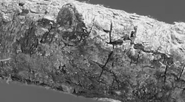}}}
   {\includegraphics[width=\textwidth]{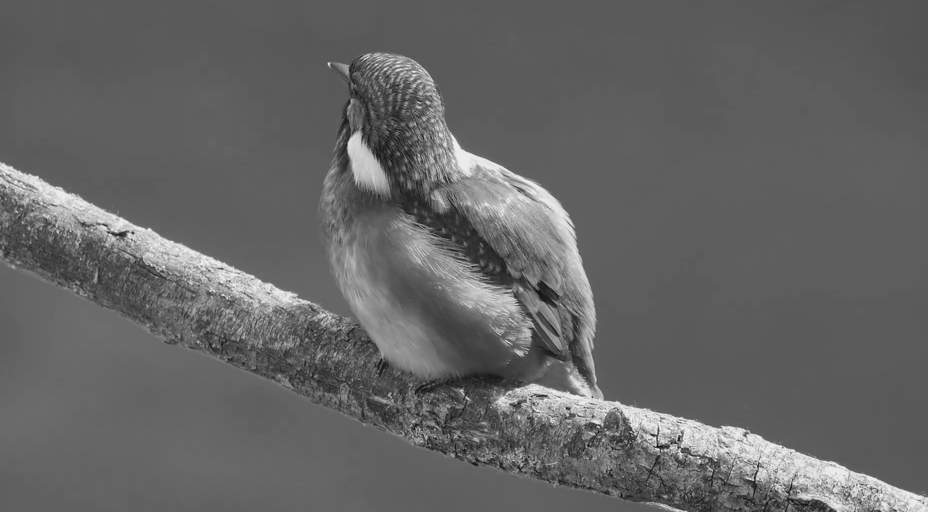}}
    \caption{GT}
    \label{fig:birdIntroGT}
\end{subfigure}  
\hfill 
\begin{subfigure}{0.475\columnwidth} 

    \includegraphics[width=\textwidth]{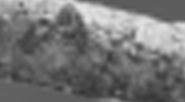}
    \caption{Bicubic}
        \label{fig:birdIntroBicubic} 
\end{subfigure}
%

\begin{subfigure}{0.475\columnwidth} 

    \includegraphics[width=\textwidth]{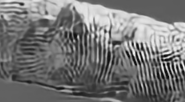}
    \caption{RLSP}
        \label{fig:birdIntroRLSP} 
\end{subfigure}
\hfill 
\begin{subfigure}{0.475\columnwidth} 

    \includegraphics[width=\textwidth]{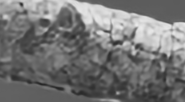}
    \caption{MRVSR (Ours)}
        \label{fig:birdIntroMRVSR} 
\end{subfigure}

\caption{A comparison between a state-of-the-art recurrent VSR network (RLSP) and our proposed network. The former generates high frequency artifacts on long sequences with low motion. The proposed network does not.}
        \label{fig:Intro} 

\end{figure}


\section{Introduction} \label{intro}

Video super-resolution (VSR) is an inverse problem that extends single-image super-resolution (SISR). While SISR aims to generate a high-resolution (HR) image from its low-resolution (LR) version, in VSR the goal is to reconstruct a sequence of HR images from the sequence of their LR counterparts. The idea behind VSR, which makes it fundamentally different from SISR, is that the fusion of several LR images produces an HR image. Therefore, VSR requires to accumulate information over a number of LR frames as large as possible. Classical VSR methods based on the image sequence formation model, knowledge on motion and iterative algorithms~\cite{farsiu2004fast,liu2011bayesian} could fill this requirement. However, these iterative algorithms are relatively slow and not suitable for real-world applications. Moreover, they perform poorly when the image sequence formation model and the assumptions on motion are too simplified.

VSR has recently benefited from DL methods~\cite{sajjadi2018frame,wang2019edvr,jo2018deep,yi2019progressive,fuoli2019efficient,isobe2020video,isobe2020videotga} that can overcome some of the drawbacks of classical methods. Deep VSR networks can efficiently learn complex spatio-temporal statistics from a training dataset of natural videos, and once trained the reconstruction is faster. There are broadly two classes of deep VSR methods. The first one groups \textbf{sliding-window based} models. These models~\cite{wang2019edvr,jo2018deep,yi2019progressive,isobe2020videotga, liu2020end} take a batch of multiple LR frames as input to fuse them and reconstruct
an HR frame. In most cases,  this batch contains 5 to 7 LR frames. Therefore, the temporal receptive field---\ie~the number of LR frames that are used in order to super-resolve a frame---is limited to 7. In contrast, methods introduced in~\cite{sajjadi2018frame,fuoli2019efficient,isobe2020video}, that build upon \textbf{recurrent} models, enable a larger temporal receptive field. In these networks, to super-resolve a frame at time step $t$, the hidden states and/or output computed in previous time step $t-1$ are taken as input, in addition to a batch of 1 to 3 LR frames. This recursion mechanism allows to propagate information through a large number of frames. As their input batch contains less LR frames and their network structures are mostly simpler, recurrent methods are faster than sliding-window based methods. Moreover, an inference of a recurrent model presents less redundant computations than the one of a sliding-window based model because each frame is processed only once. Finally, sliding-window based VSR methods generate independent output HR frames, which reduces temporal consistency of the produced HR frames, resulting in flickering artifacts. This is not the case for recurrent VSR, in which information about previously super-resolved frame is part of the input at each time step. These considerations make recurrent methods more interesting from a realistic application-oriented point of view.


Because of computational and memory constraints, as well as vanishing and exploding gradients, recurrent VSR models are usually trained on sequences of 7 to 12 images. They are then deployed to super-resolve a sequence of any length. Some applications, such as video-surveillance, would require to super-resolve sequences of arbitrary length. However, recurrent models are not trained on these long sequences. Hence, there is no guarantee that they optimally perform on long sequences. 
In this study, we show that recurrent VSR networks generate high frequency artifacts when inferring on long video sequences presenting low motion. Such sequences contain parts of the scene that barely move, for instance when the camera is quasi-static. 
The super-resolution process creates high-frequency information which is accumulated in the long-term recurrence, creating artifacts and causing divergence.
\cref{fig:Intro} illustrates this phenomenon.
To the best of our knowledge, this work is the first study about VSR that raises this instability issue. This unexpected behavior can be critical for some real-world applications, like video surveillance in which both the camera and the scene stay static for a long time. 

The structure of the article is the following. First, we review studies related to VSR and instabilities of recurrent networks. Then, based on Lipschitz stability theory, we propose a new framework of recurrent VSR network that is both stable and competitive on long sequences with low motion. After this, we introduce a new recurrent VSR network MRVSR as an implementation of this framework. Finally, we empirically analyze instabilities of existing recurrent VSR models on long sequences with low motion and show the stability and superior performance of the proposed network. A new long sequence dataset has been created for our experiments. We will make it publicly available.

\section{Related work}

\subsection{Recurrent video super-resolution}

Authors of~\cite{sajjadi2018frame} were pioneers of recurrent VSR. They introduced FRVSR, in which the previous output frame is warped based on a dense optical flow estimation and fed back as an additional input to a super-resolution network at the next time step. The optical flow is estimated by another network and the two networks are jointly trained end-to-end. Hence, FRVSR operates \textit{frame-recurrence}.

A more recent recurrent VSR architecture called \textit{recurrent latent space propagation} (RLSP) was introduced in~\cite{fuoli2019efficient}. In this approach, the previous output frame and the previously estimated locality based hidden state are used as an extra input at the next time step. Compared to frame-recurrence, RLSP can be interpreted as maximizing the depth and width of the recurrent connection. 
In contrast to FRVSR, RLSP is based on implicit motion compensation. The overall architecture is computationally efficient, which enables RLSP to be the fastest VSR network at this time. 

RSDN~\cite{isobe2020video} is so far the recurrent VSR network that reportedly performs the best for relatively short sequences, according to its performance on Vid4 dataset, composed of 4 videos between 34 to 49 frames~\cite{liu2011bayesian}. Its architecture presents a recurrent hidden state coupled with a hidden-state adaptation module and structure-detail decomposition. The input LR frames and the hidden state are decomposed into structure and detail components and fed to two interleaved branches to reconstruct the corresponding components of HR frames.

\subsection{Instabilities of recurrent neural networks}

Recurrent Neural Networks (RNNs) are difficult to train~\cite{pascanu2013difficulty}. First of all, they involve backpropagation through time (BPTT), \ie~their unrolling through time, that is costly in terms of memory. Secondly, these architectures risk vanishing and exploding gradients issues. Correlated to this, RNNs are prone to divergence when inferring on long sequences. Authors of~\cite{miller2018stable} showed, in the context of multi-layer and LSTM networks, that an RNN is stable if its Lipschitz constant is smaller than 1
. To enforce this constraint, they proposed to clip singular values of the matrix associated with the recurrence map to 1. Several works circumvent vanishing and exploding gradients problems by setting all the singular values to 1~\cite{arjovsky2016unitary,wisdom2016full,mhammedi2017efficient,vorontsov2017orthogonality,jose2018kronecker,zhang2018stabilizing}. 

Some studies are related to enforcing the Lipschitz constraint in the context of convolutional neural networks. Authors of~\cite{sedghi2018singular} proposed to clip singular values of the block matrix of doubly block-circulant matrices associated with the convolutional layer. 
The work~\cite{miyato2018spectral} explored \textit{spectral normalization}, that relies on the power iteration to estimate
maximal singular value of the reshaped kernel tensor of the convolutional layer. Authors of~\cite{scaman2018Lipschitz,gouk2021regularisation} suggested not using this reshaping and instead proposed to directly use the kernel tensor in the power iteration. Finally, the work~\cite{sanyal2019stable} proposed Stable Rank
Normalization (SRN), an algorithm that seeks to enforce either the Lipschitz constraint or its softer version. 

In the context of recurrent video denoising, authors of~\cite{tanay2020diagnosing} pointed out instabilities. They first brought out unforeseeable, colorful and black mask-like artifacts in long-term video denoising. 
Then, inspired by studies on adversarial examples~\cite{goodfellow2014explaining}, they proposed a diagnosis tool to check stability of a trained recurrent video processing network. 
Finally, they improved upon the SRN algorithm to propose \textit{Stable Rank Normalization of Layer} (SRNL). While SRN reshapes the kernel tensor of the convolutional layer, SRNL avoids this reshaping, similarly to~\cite{scaman2018Lipschitz,gouk2021regularisation}. They applied this method on convolutional layers of their recurrent video denoising network and demonstrated its effectiveness. 


To conclude this section, the following points summarize the limits of existing works regarding long-term recurrent VSR and our contributions: 

\begin{itemize}

\item existing recurrent VSR networks have been only evaluated on relatively short generic sequences. Their performances have not been measured on long sequences. We demonstrate these networks perform poorly on such sequences when the motion amplitude is low, due to their recurrent structure. We create a novel dataset of long and low motion sequences, because existing datasets only contain sequences that either are too short or present fast scene motion;

\item the relationship between instabilities and scene motion in video has not been investigated. 
We show that when inferring on long sequences presenting low motion, existing recurrent VSR models diverge;

\item the Lipschitz constraint has not been applied on existing recurrent VSR networks. Indeed, in order to have a stable recurrent VSR network, we could first take one of these networks and directly apply a Lipschitz constraint to all convolutional layers in the recurrent loop. We show that this strategy fails when super-resolving long sequences with low motion;

\item we design a recurrent VSR framework that is stable on long sequences with low motion, while not being globally Lipschitz constrained. We demonstrate the superior performance of a network based on this framework. 


\end{itemize}

\section{Method}

\subsection{Stability of recurrent video processing models} \label{srnl}

A recurrent video processing model is determined by a \textit{recurrence map} $\phi  ^ L : \mathbb{R}^n \times \mathbb{R}^d \rightarrow \mathbb{R}^n $ and an \textit{output map} $\psi  : \mathbb{R}^n \rightarrow \mathbb{R}^c $. The recurrent information $h_t \in \mathbb{R}^n $ and the output image $\hat{y}_t \in \mathbb{R}^c $ are updated at each time step $t$ as follows:
\begin{equation}
 \begin{cases}
 h_t = \phi ^L (h_{t-1}, x_t) \\
 \hat{y}_t = \psi (h_{t})
 \end{cases}
 \label{vpm}
\end{equation}
\noindent where $x_t \in [0,1]^d $ is an input image provided at time $t$.

%
The recurrent model is \textit{Lipschitz stable} if $\phi ^L $ is \textit{contractive} in $h$ \ie~if $\phi ^L $ is $L$-Lipschitz in $h$ with $L < 1$ (the superscript in $\phi   ^L$ highlights this Lipschitz continuity). $L$ is the Lipschitz constant of $\phi   ^L$. 
This stability ensures that the full recurrent system is globally stable when running the network an arbitrary number of times, avoiding any divergence.
Assume that $\phi  ^ L$ is composed of $K$ convolutional layers 
interspaced with ReLU non-linearities. Each convolutional layer can be encoded by a weight matrix, obtained from the layer’s kernel tensor as a block matrix of doubly block-circulant matrices. Because Lipschitz constant of the ReLU activation is 1, $L$ is upper-bounded by the product of the spectral
norms of the weight matrices of the convolutional layers:

\begin{proposition} 
For a recurrent model $\phi  ^L $ constituted of $K$ convolutional layers with weight matrices $W_1, ...,W_K \in \mathbb{R} ^ {n \times n}$ interspaced with ReLU non-linearities, the Lipschitz constant $L$ of $\phi  ^L $ verifies:~\begin{equation}
L \leq \prod_{k=1}^{K} {||W_k||}
\label{upper}
\end{equation}~where $ ||.|| $ is the spectral norm.
\end{proposition}

Given this inequality, the Lipschitz stability can be ensured under the hard Lipschitz constraint:


\begin{constraint} \textbf{Hard Lipschitz constraint (HL)}

\noindent 
$\forall k \in [\![ 1, K ]\!] $, we impose $ || W_k || \leq 1 $.
\end{constraint}

However, the upper bound in \cref{upper} mostly overestimates $L$. For example, if $\phi ^L$ is constituted of 2 convolutional layers with weight matrices $W_1$ and $W_2$, the only case where $L = ||W_1|| \cdot ||W_2|| $ is when the first right singular vector of $||W_1||$ and the first left singular vector of $W_2$ are aligned. Hence, the constraint is overly restrictive. One can thus decide to relax the latter, leading to the soft Lipschitz constraint:

 
\begin{constraint} \textbf{Soft Lipschitz constraint (SL)}

\noindent $\forall k \in [\![ 1, K ]\!]$, we set $ || W_k || = \alpha > 1 $ and minimize srank($W_k$) based on training data, where srank is the Stable rank.
\end{constraint}

 \textit{Stable rank} is an approximation of the rank operator that is stable under small perturbations of the matrix. This soft constraint does not theoretically guarantee the Lipschitz stability, so it is important to empirically verify the non divergence.

To enforce these constraints in the context of convolutional neural networks, \textit{Stable Rank Normalization of Layer} (SRNL) can be applied to a convolutional layer during the training stage. This sets the spectral norm of the matrix of this layer to a desired value $\alpha$ and minimizes the stable rank of the matrix during training, controlled by $\beta$. $\alpha$ and $\beta$ are among hyperparameters of the algorithm. When $\beta = 1$, it is equivalent to performing spectral normalization on the matrix. 
After training, a normalization step is required just before test time, so the algorithm does not introduce any overhead in inference speed. 

\subsection{Unconstrained Stable Recurrent VSR framework} \label{USRVSR}


In approaches such as RLSP, FRVSR and RSDN, every convolutional layer of super-resolving networks is recurrent within feedback loops. This seeks to increase the depth and width of the recurrent connection by giving the hidden state and the previous output to the input of super-resolving networks. Therefore, these layers both incorporate past information and contribute to the deconvolution task. Adopting the notations from \cref{vpm}, in these networks $\psi$ is reduced to the identity mapping (followed by pixel shuffling or transposed convolutions). In order to have a stable recurrent VSR network, a naive approach would be to directly apply SRNL to one of these VSR networks. However, this approach presents some difficulties. 

First, we applied SRNL to RLSP with hyperparameters $(\alpha, \beta) = (2.0, 0.1)$ and empirically verified that SL 
was not capable of removing the artifacts on long sequences (see \cref{fig:birdRLSPSRNL201}). Second, we did the same experiment with $(\alpha, \beta) = (1.0, 1.0)$ to enforce HL and this resulted in a stable network but with poor VSR performance (detailed in \cref{resultsStab}). This is because the resulting architecture has been constrained to be globally 1-Lipschitz, and a successful super-resolving function---that operates both upsampling and deconvolution---can not be 1-Lipschitz; since some frequencies need to be boosted as the Wiener filter does in the optimal linear case. This is not the case for a denoising function, that can be 1-Lipschitz while correctly performing. 

Considering these points, we define a new framework of recurrent VSR network that is stable and performs competitively on long sequences:


%


\begin{definition}
An \textbf{Unconstrained Stable Recurrent VSR} network is defined by an input network $\xi  : [0,1]^{d \times (2T+1)} \rightarrow \mathbb{R}^d $, a contractive recurrent network $\phi  ^L : \mathbb{R}^n \times \mathbb{R}^d \rightarrow \mathbb{R}^n $ and an output network $\psi   : \mathbb{R}^n \rightarrow \mathbb{R}^c $. The features $z_t$, the hidden state $h_t$ and the output image $\hat{y}_t$ are updated at each time step $t$ as follows:
\begin{equation}
 \begin{cases}
  z_t = \xi (X_t) \\
 h_t = \phi ^L (h_{t-1}, z_t) \\
 \hat{y}_t = \psi (h_{t})
 \end{cases}
\end{equation}
\noindent where $X_t = \{ x_{t} \} _{ t - T \leq t \leq t + T } \in [0,1]^{d \times (2T+1)} $ is an input batch of LR images provided to the network at $t$ and $2T + 1$ denotes the size of the batch.

Let $\phi  ^L $ be constituted of $K$ convolutional layers with weight matrices $W_1, ...,W_K \in \mathbb{R} ^ {n \times n}$ interspaced with ReLU activations. $\phi  ^L $ is contractive in $h$ based on the hard Lipschitz constraint: $\forall k \in [\![ 1, K ]\!] $, $ || W_k || \leq 1 $. 
\end{definition}

\par \textbf{Stable}: all the layers in the inner recurrent loop of such a network are contractive, which guarantees its stability over time. 

\par \textbf{Unconstrained}: such a network is not globally constrained in terms of Lipschitz continuity, due to its non contractive input and output networks which can keep their full expressiveness. 

Most of the deconvolution task is done by $\xi$ and $\psi$. $\phi ^L$ incorporates past information. When $\xi$ and $\psi$ are simultaneously identity mappings, the \textit{unconstrained} property is lost, as the network becomes globally 1-Lipschitz. This is the case encountered when imposing HL on all convolutional layers of networks such as RLSP, FRVSR and RSDN.


\subsection{Middle Recurrent Video Super-Resolution} \label{MRVSR}

\begin{figure}[!htbp]
\centering
   \includegraphics[width=1\linewidth]{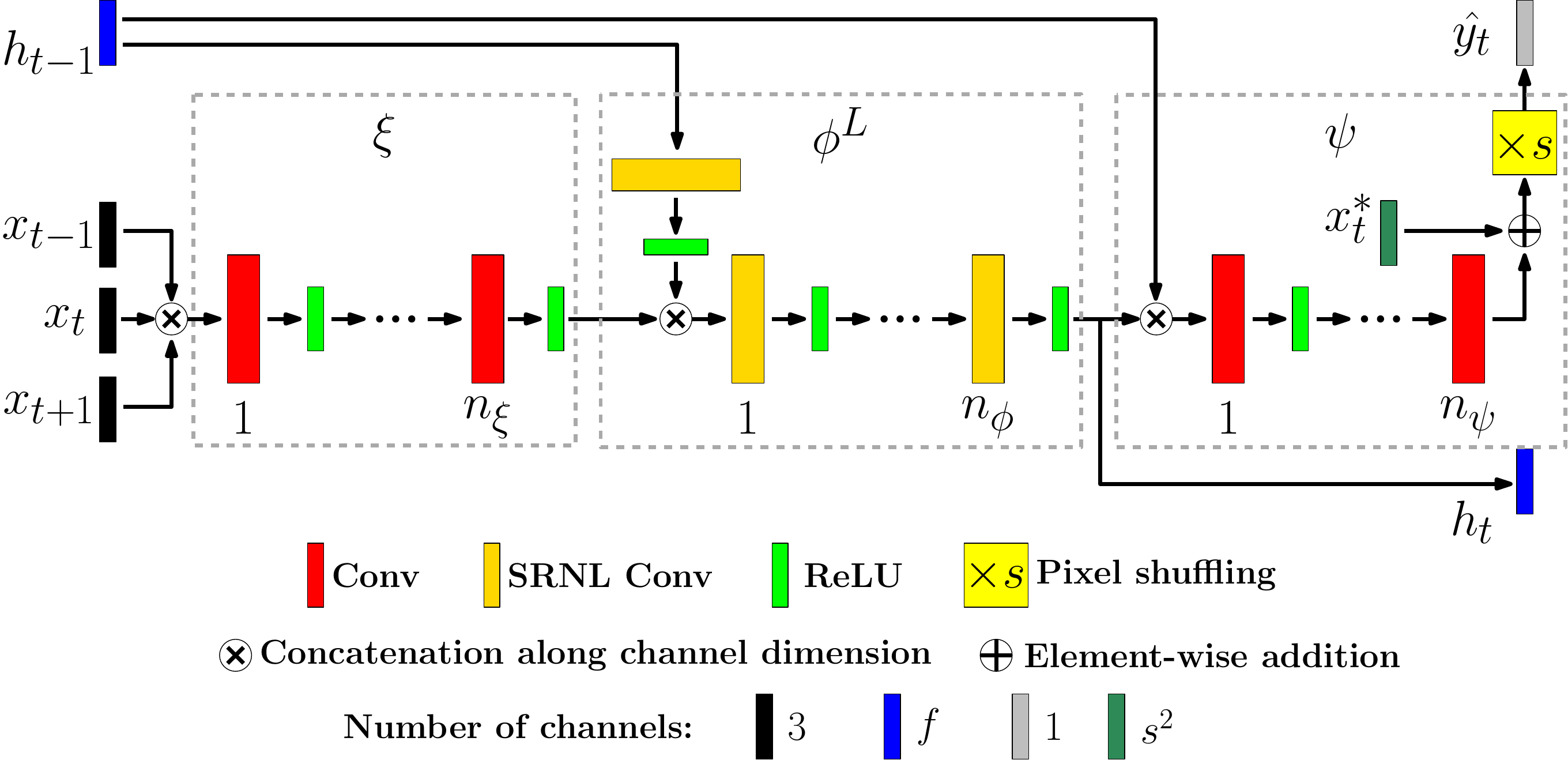}
\caption{MRVSR architecture. SRNL Conv denotes convolutional layer under HL enforced by SRNL. Each convolutional layer uses $3 \times 3$ kernel with stride 1 and outputs $f$ feature maps ($ f =128$ in our study), except the last one which outputs $s^2 = 16$ feature maps, where $s$ is the scaling factor. The network outputs the brightness channel Y of YCbCr color space. Cb and Cr channels are upsampled independently with bicubic interpolation. Input LR frames $\{x_i\}_{t-1 \leq i \leq t + 1}$ are in RGB colorspace. Besides, $x_t$ is converted from RGB to Y and replicated $s^2 = 16$ times in the channel dimension, which gives $x^{\star}_t$ for the residual connection. Pixel shuffling rearranges elements in a tensor of shape $(C \times s^2, H, W)$ to a tensor of shape $(C, H \times s, W \times s)$.
}
\label{fig:MRVSR}
\end{figure}

As an implementation of the proposed framework, we design a new network coined \textbf{Middle Recurrent Video Super-Resolution} (MRVSR). Its architecture is illustrated in \cref{fig:MRVSR}. 
The first part of the network, $\xi$, has a feed-forward architecture with $n_\xi$ convolutional layers and interspaced ReLU activations.  
The second part $\phi ^L$ is composed of $n_\phi + 1$ convolutional layers under HL and interspaced ReLU activations. 
The third part $\psi $ has a feed-forward architecture with $n_\psi$ convolutional layers interlaced with ReLU activations and followed by a pixel shuffling layer. This part takes as input the current hidden state $h_t$ and the hidden state from the previous timestep. This mecanism, called \textit{feature-shifting}, 
is helpful to promote temporal consistency between two successively output frames. 

Incorporating past information via the recurrent connection is a simpler task than deconvolution. This can be illustrated revisiting a traditional non DL based VSR method as an example: in the classical Shift-and-Add agorithm~\cite{farsiu2004fast}, historical information is captured via averaging or median aggregating past frames after projection on a HR grid and motion compensation. Averaging or median aggregating are rather simple mathematical operations. Therefore, $ n_\phi $ can be smaller than $n_\xi + n_\psi$. In practice, one can fix $ n_\xi + n_\phi + n_\psi $ to satisfy some constraint on computational cost, set a small value for $n_\phi$ and then select $n_\xi$ and $n_\psi$. In our setting, we have found that under the condition $ n_\xi + n_\phi + n_\psi = 7 $ (that enables both fast computations and good performance), the value $n_\phi = 1$ lead to the best performance among other values of $n_\phi$ on our validation set (described in \cref{datasets}).



\section{Experiments} \label{experiments}


\subsection{Networks}\label{networks}

For comparison, we implement the following state-of-the-art recurrent VSR networks in Pytorch~\cite{paszke2017automatic}: FRVSR 10-128~\cite{sajjadi2018frame}, RSDN 9-128~\cite{isobe2020video} and RLSP 7-128~\cite{fuoli2019efficient}. The numbers after each network respectively indicate the number of repeated building blocks and the number of filters in each convolutional layer. These hyperparameters enable reasonably fast training and testing and satisfactory performance on short sequences. In the following, we omit these numbers for simplicity. For RSDN, our implementation is based on the official codes released by its authors.\footnote{\url{https://github.com/junpan19/RSDN}} Additionally, we implement modified RLSP where all its layers have been normalized by SRNL with hyperparameter sets $(\alpha, \beta) =(2.0, 0.1) $ and $(\alpha, \beta) =(1.0, 1.0) $ to enforce the soft and hard Lipschitz constraints respectively. 
We call these networks RLSP-SL and RLSP HL.


We compare these networks against the proposed MRVSR. 
We select $ (n_\xi, n_\phi, n_\psi)$ so that $n_\xi+n_\phi+n_\psi=7$ for the reason stated in \cref{MRVSR}. This number equals the number of convolutional layers in RLSP (excluding the layer that processes the hidden state), which yields fair comparison. Among MRVSR with different sets $ (n_\xi, n_\phi, n_\psi)$, the network with $ (n_\xi, n_\phi, n_\psi) = (3,1,3) $ was the best performing model on our validation set. Therefore, in \cref{results} we only report performances recorded by MRVSR with this hyperparameter set. We use SRNL with $(\alpha, \beta) =(1.0, 1.0) $ to impose the HL. 

In order to measure the benefit from constrained recurrence map, we also implement MRVSR without its recurrence and feature-shifting, which coincides with RLSP without its recurrence mecanism. This can be seen as an extension of SISR that takes 3 consecutive LR frames as an input at each timestep. Its architecture is feed-forward with 7 convolutional layers with interlaced ReLU activations. We call this network RFS3 for \textbf{R}esidual \textbf{F}usion \textbf{S}huffle network with \textbf{3} input frames. This network will serve as baseline against recurrent models. In addition, we also implement RFS with an input batch of 7 LR frames, that we call RFS7. This serves as a representative sliding-window based model to compare against MRVSR, because most of sliding-window based VSR models take a batch of 5 to 7 LR frames.



\subsection{Datasets}\label{datasets}

We prepare the training dataset in a similar way as in~\cite{fuoli2019efficient}. From the 37 high resolution Vimeo
videos that were used in this study, after downsampling them by a factor of 2 we extract 40,000 random cropped sequences of size
$I \times 256 \times 256 \times 3 $, where $I \geq 12$. The delimiting keyframes are excluded from the sequence. At training time, we sample random sub-sequences of these crops with length $ 12 $. By excluding the first and the last frames, we obtain ground truth (GT) sequences with length $10$. The first and last frames of the sampled sequences are used to produce $ x_{-1} $ at the beginning and $ x_{10} $ at the end. Data augmentation (random flip/transposition) is also employed.


We also prepare a validation set of 4 sequences. They come from videos with no constraints on motions of objects and count between 30 and 50 frames each.

We introduce a new test set of long sequences in which the camera is quasi-static and foreground objects move. This dataset will be complementary to existing datasets (Vid4~\cite{liu2011bayesian}, REDS~\cite{Nah_2019_CVPR_Workshops_REDS} and Vimeo-90K~\cite{xue2019video}) which contain only videos that either are short, or present fast scene motion. To generate this new dataset, we download videos from \texttt{vimeo.com} and \texttt{youtube.com} and extract 4 sequences with quasi-static scene and moving objects inside. The first two of them are respectively Full HD and HD Ready and the two others are 4K. The HD and 4K sequences are downsampled respectively by a factor of 2 and 4. These 4 sequences respectively have the following lengths in number of frames: 379, 379, 379 and 172. They constitute the test dataset we call \textbf{Quasi-Static Video Set}. We limited the lengths of the sequences to 379 to ensure dataset homogeneity, but the video containing the first sequence contains a much larger number of frames. Therefore, we have also prepared a longer version of the first sequence called \textit{Sequence 1-XL}. The latter contains 8782 frames. We will make all of these sequences available on \url{https://github.com/bjmch/StableVSR}.


The train and validation sets contain standard, relatively short sequences with no constraints on motion, whereas the test set contains long sequences with low motion. It aims at testing the capability of networks trained on short sequences to work on real-life long sequences that may have low-motion periods. We remind the reader that training recurrent networks on such long sequences is not realistic for reasons explained in \cref{intro}, so the generalization gap between short and long sequences can not be addressed with training data.

We additionally compare the reconstruction performances on the standard Vid4 dataset.

From each of the training, validation and test sequences in HR space, the corresponding LR sequence is generated by applying gaussian blur with $ \sigma$ and sampling every $s = 4$ pixel in both spatial dimensions. 
We set $ \sigma = 1.5$, except when testing RSDN. In the case of this network, we use the pre-trained weights available on its official github repository. We thus adapted the codes of the corresponding degradations that are available on this repository to generate the LR sequence and the value of $ \sigma = 1.6$ was used.

\subsection{Training procedure}
All of the networks we prepare are trained from scratch after the Xavier initialization~\cite{glorot2010understanding}, except RSDN. The loss function is pixel-wise mean-squared-error between pixels in the brightness channel Y of YCbCr color space of GT frames and the network’s output. The networks are trained with Adam optimizer~\cite{kingma2014adam} and a batch size of 4. The learning rate starts at $10^{-4}$ and is divided by 10 after the 200th and 400th epochs. RFS3, RFS7 and MRVSR are trained for 600 epochs. Other models except RSDN are trained between 400 and 600 epochs until convergence, based on train and validation losses.


\subsection{Evaluation}

We numerically evaluate the networks based on frame PSNR and SSIM. Qualitative evaluation that checks the presence of artifacts is of equal importance. We also assess the temporal consistency by examining temporal profiles from output sequences. 

Moreover, the diagnosis tool from~\cite{tanay2020diagnosing} can be used in order to visualize Spatio-Temporal Receptive Field (STRF) of a recurrent network. This tool, that is inspired by studies on adversarial examples~\cite{goodfellow2014explaining}, works as follows: given a trained recurrent video processing network, it looks for an input sequence $ X = (x_{-\tau}, ..., x_{\tau}) $ that is  optimized  to  trigger instabilities in the output sequence $ Y = (y_{-\tau +1}, ..., y_{\tau - 1}) $.  To do so, the L1 norm of the center pixel $|p|$ in $y_0$ is maximized. This optimization only affects pixels in $X$ that have an effect on the value of $p$. Therefore, the optimized sequence $X$ can be interpreted as a  visualization of the STRF for the pixel $p$. $\tau$ is typically set to 40, pixels in $ X $ are randomly initialized following an uniform random variable between 0 and 1 and images in $X$ have dimensions $ 64 \times 64 \times 3 $. In our experiment, the optimization is solved using gradient descent and Adam optimizer for 1500 iterations. The learning rate starts at 1 and is divided by 10 after 750 and 1250 iterations. 

\section{Results} \label{results}

\subsection{Performance of existing recurrent networks}

\begin{figure}[!htbp]

\centering
   \includegraphics[width=1\linewidth]{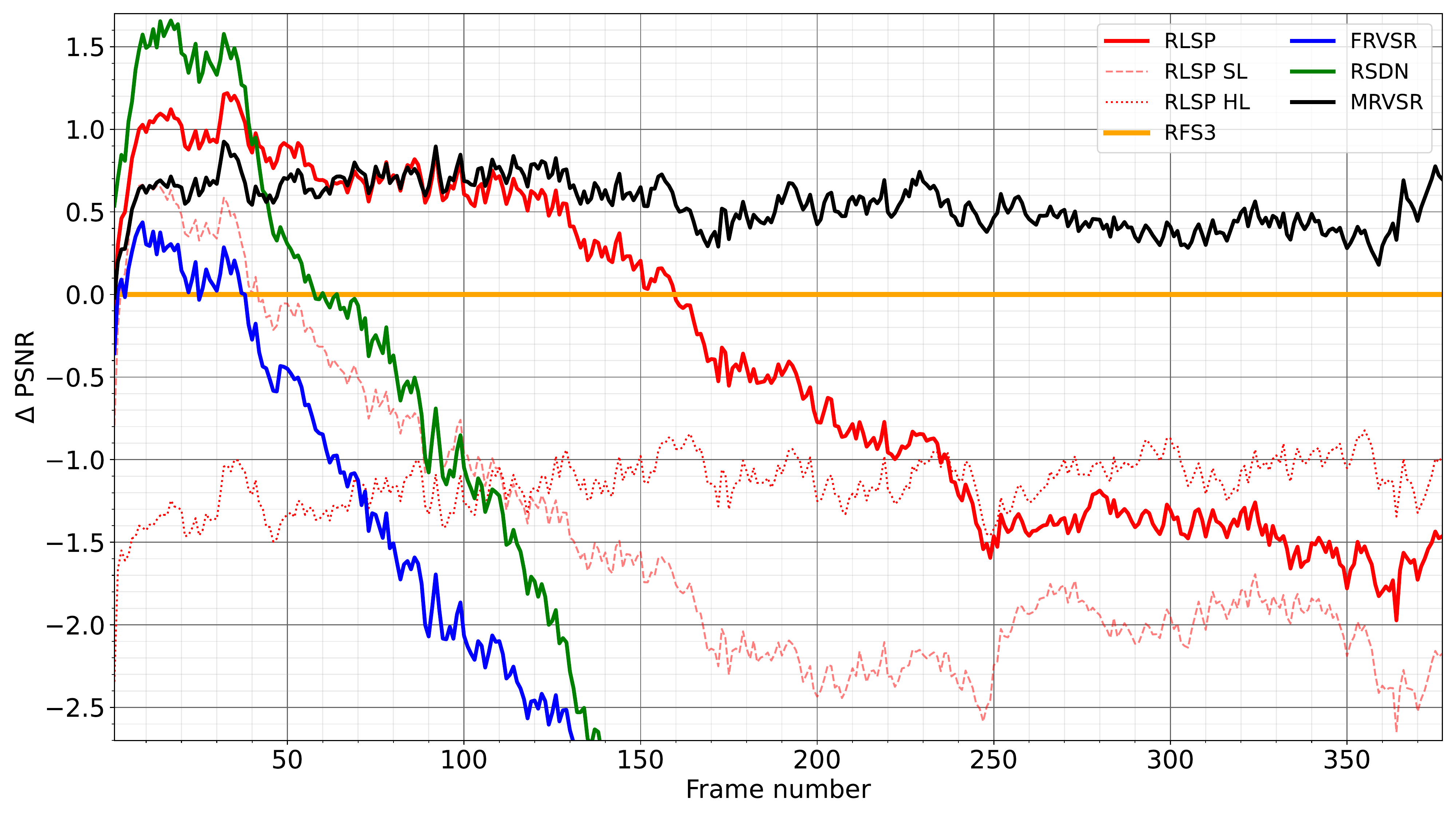}
\caption{Evolution of PSNR on Y channel per frame averaged over the first three sequences of the Quasi-Static Video Set. We substract the curve of the RFS3 baseline and the graph shows these differences.
}
\label{fig:recurrentnets}
\end{figure}

\cref{fig:recurrentnets} shows the evolution of the PSNR per frame for some of the networks, averaged over the first three sequences of Quasi-Static Video Set. The curve of RFS3 is taken as a baseline and subtracted to the other ones, and the resulting curves are displayed. We see that until a relatively small number of processed frames, existing recurrent networks (RLSP, RSDN and FRVSR) perform optimally and remain better than the baseline model. 
But at a certain point their performance drop and they become worse than the baseline model, indicating that the recursion integrates harmful information at each new frame. This can be seen as divergence.

\begin{table}[h]

        \centering
        \setlength\tabcolsep{1.75pt}
        \begin{tabular}{| l | l | l | p{2cm} |}
        \hline
        Model & First 50 & All & Last 50 \\
        \hline
        Bicubic & 30.08 / 0.8362 & 30.05 / 0.8356 & 30.11 / 0.8387\\
        RFS3 & 32.20 / 0.8911 & 32.04 / 0.8886 & 32.07 / 0.8911\\
        RFS7 & 32.38 / 0.8945 & \textcolor{blue}{32.23} / 0.8921 & \textcolor{blue}{32.26} / \textcolor{blue}{0.8943}\\

        FRVSR & 32.15 / 0.8947 & 29.16 / 0.8442 & 27.68 / 0.8121\\
        RSDN & \textcolor{red}{33.46} / \textcolor{red}{0.9181} & 29.82 / 0.8788 & 27.98 / 0.8549\\
        RLSP & \textcolor{blue}{33.08} / \textcolor{blue}{0.9099} & 31.67 / \textcolor{blue}{0.8964}& 30.57 / 0.8882\\
        RLSP-SL & 32.45 / 0.8991 & 30.62 / 0.8708 & 29.98 / 0.8627\\
        RLSP-HL & 30.98 / 0.8618 & 30.91 / 0.8608 & 30.95 / 0.8630\\
        MRVSR & 32.80 / 0.9030 & \textcolor{red}{32.62} / \textcolor{red}{0.9007}  & \textcolor{red}{32.62} / \textcolor{red}{0.9026} \\     
        \hline           
       \end{tabular}
       \caption{Mean PSNR / SSIM on Y channel of Quasi-Static Video Set. The metrics are measured excluding the first 3 and last 3 GT frames. `First 50' means the metrics are computed at the beginning of the sequences \ie~on the first 50 reconstructed frames. `All' means the metrics are computed through the entire sequences \ie~on all reconstructed frames. `Last 50' means the metrics are computed at the end of the sequences \ie~on the last 50 reconstructed frames. \textcolor{red}{Red}: the best result. \textcolor{blue}{Blue}: the second best result.}
       \label{tab:psnrssim}

\end{table}

\begin{figure}[!htbp]

\centering

\begin{subfigure}{0.475\columnwidth} 
\stackinset{r}{-.\textwidth}{t}{-.\textwidth}
   {
   \setlength{\fboxsep}{0pt}%
 \setlength{\fboxrule}{1pt}%
   \fbox{\includegraphics[width=0.4\textwidth]{Bird/RLSP_375_cropped_artifact.png}}}
   {\includegraphics[width=\textwidth]{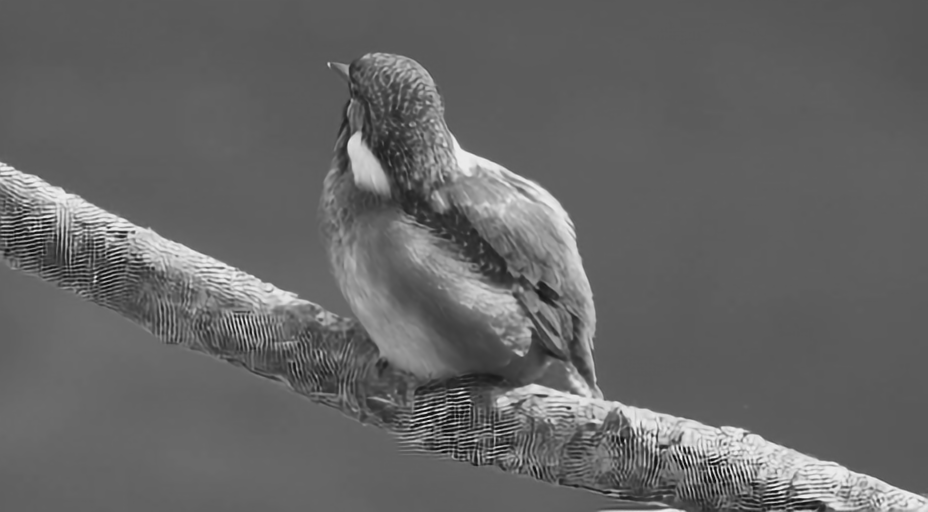}}
    \caption{RLSP}
    \label{fig:birdRLSP}
\end{subfigure}  
\hfill 
\begin{subfigure}{0.475\columnwidth} 
\stackinset{r}{-.\textwidth}{t}{-.\textwidth}
   {
   \setlength{\fboxsep}{0pt}%
 \setlength{\fboxrule}{1pt}%
   \fbox{\includegraphics[width=0.4\textwidth]{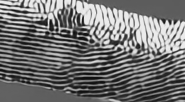}}}
    {\includegraphics[width=\textwidth]{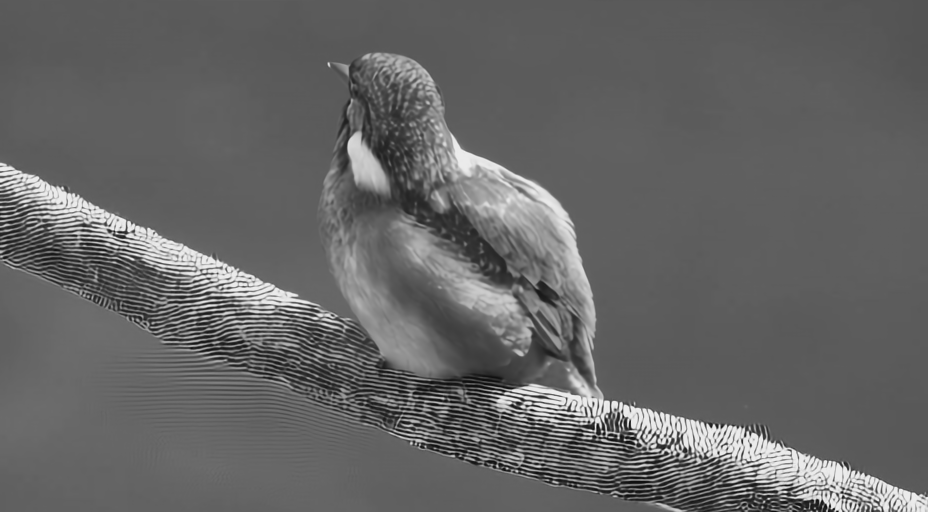}}
    \caption{RSDN}
        \label{fig:birdRSDN} 
\end{subfigure}
%

\begin{subfigure}{0.475\columnwidth} 
\stackinset{r}{-.\textwidth}{t}{-.\textwidth}
   {
   \setlength{\fboxsep}{0pt}%
 \setlength{\fboxrule}{1pt}%
   \fbox{\includegraphics[width=0.4\textwidth]{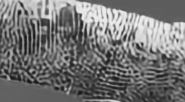}}}
   {\includegraphics[width=\textwidth]{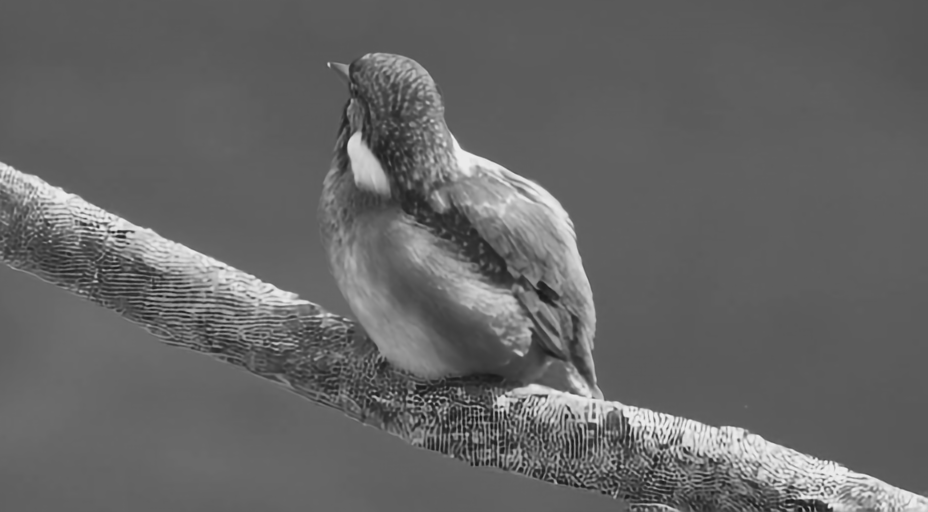}}
    \caption{FRVSR}
        \label{fig:birdFRVSR} 
\end{subfigure}
\hfill 
\begin{subfigure}{0.475\columnwidth} 
\stackinset{r}{-.\textwidth}{t}{-.\textwidth}
   {
   \setlength{\fboxsep}{0pt}%
 \setlength{\fboxrule}{1pt}%
   \fbox{\includegraphics[width=0.4\textwidth]{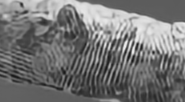}}}
   {\includegraphics[width=\textwidth]{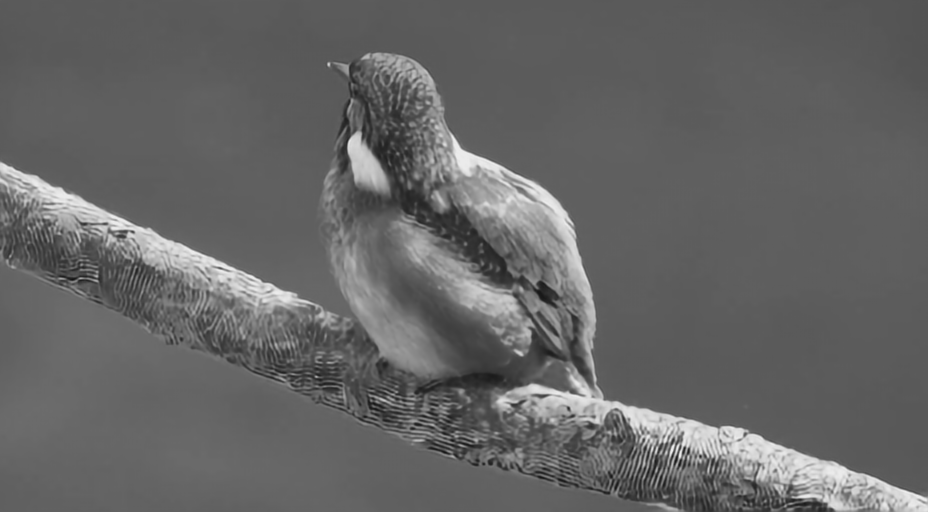}}
    \caption{RLSP-SL}
    \label{fig:birdRLSPSRNL201} 
\end{subfigure}  

\caption{A frame near the end of the first sequence of Quasi-Static Video Set (the 376th frame) reconstructed from state-of-the art recurrent networks, and RLSP-SL. The Y channel is visualized. 
The networks generate high frequency artifacts on the branch, which is a quasi-static object.} 

\label{fig:bird}
\end{figure}

%







\cref{tab:psnrssim} summarizes the performances of the networks on the Quasi-Static Video Set. It summarizes the performances of the methods at the beginning of the sequences, through the entire sequences, and at the end of the sequences. The table conforms with the curves shown on \cref{fig:recurrentnets}. Based on reported performances, at the beginning of the sequences RLSP and RSDN perform better than the baseline RFS3. 
However, at the end of the sequences these networks and FRVSR have diverged and perform worse than RFS3. The differences in performance on the last 50 reconstructed frames between RFS3 and respectively RLSP, FRVSR and RSDN are $-1.50$, $-4.39$ and $-4.09$ in PSNR and $-0.0029$, $-0.0790$ and $-0.0362$ in SSIM. They represent in average $-3.33$dB in PSNR and $-0.0394$ in SSIM. This performance drop is due to the generation and accumulation of high frequency artifacts. These artifacts appear on objects that barely move. 
Example artifacts are shown on \cref{fig:birdRLSP,fig:birdRSDN,fig:birdFRVSR} which show a frame near the end of the first sequence of Quasi-Static Video Set (the 376th frame) reconstructed by each network.

\par \textbf{Behavior analysis:} These existing recurrent networks are trained to optimize their performance on a very low number of frames (at most 10). In this setting, it is beneficial to the network to produce rapidly a huge amount of details in the output sequence. These high frequency details grow in strength with time, but they are not fed back into the network more than 10 times, so the optimization process is not trained to manage their increase after this period. When inferring on long sequences, these details keep accumulating long after the short-term network's training regime, which produces visible artifacts that diverge over time.
In the presence of strong motion, even with short-term training, the network learns to forget the past information, which is inconsistent with the new one. The newly created high frequency content is forgotten at the same time, preventing divergence on scenes with enough motion. In the first sequence of the Quasi-Static Video Set, the bird moves regularly, which is why artifacts do not have time to appear on the bird itself, as can be seen on \cref{fig:bird}.

\subsection{Constraining existing recurrent networks} \label{resultsStab}

\begin{figure}[!htbp]

\centering

\begin{subfigure}{0.475\columnwidth}
 \stackinset{r}{-.\textwidth}{t}{-.\textwidth}
   {
   \setlength{\fboxsep}{0pt}%
 \setlength{\fboxrule}{1pt}%
   \fbox{\includegraphics[width=0.4\textwidth]{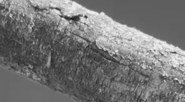}}}
    {\stackinset{l}{-2.5pt}{t}{0pt}
   {
   \setlength{\fboxsep}{0pt}%
 \setlength{\fboxrule}{1pt}%
   \fbox{\includegraphics[width=0.35\textwidth]{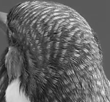}}}
   {\includegraphics[width=\textwidth]{Bird/GT_375.png}}}
    \caption{GT}
    \label{fig:GTbird}
\end{subfigure}
\hfill
\begin{subfigure}{0.475\columnwidth}
 \stackinset{r}{-.\textwidth}{t}{-.\textwidth}
   {
   \setlength{\fboxsep}{0pt}%
 \setlength{\fboxrule}{1pt}%
   \fbox{\includegraphics[width=0.4\textwidth]{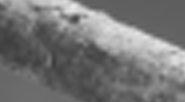}}}
    {\stackinset{l}{-2.5pt}{t}{0pt}
   {
   \setlength{\fboxsep}{0pt}%
 \setlength{\fboxrule}{1pt}%
   \fbox{\includegraphics[width=0.35\textwidth]{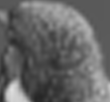}}}
   {\includegraphics[width=\textwidth]{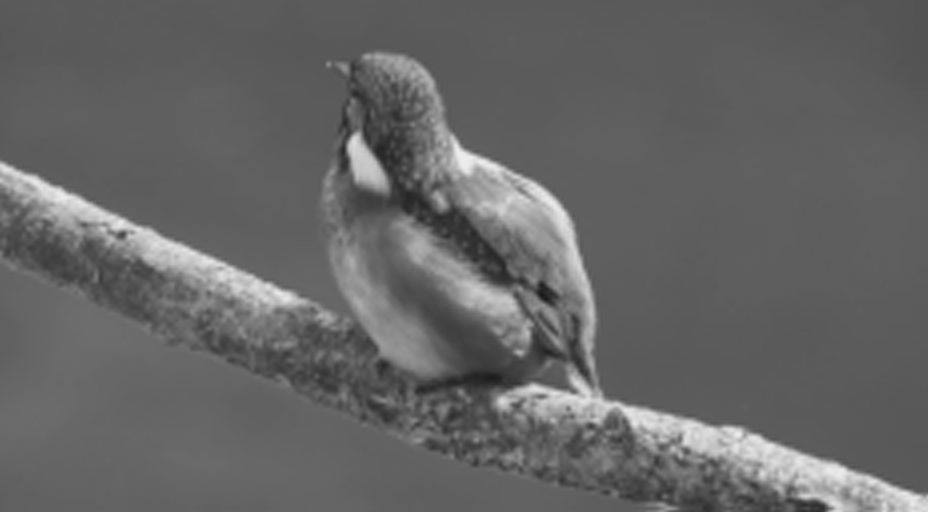}}}
    \caption{Bicubic}
    \label{fig:Bicubicbird}
\end{subfigure}
\begin{subfigure}{0.475\columnwidth}
 \stackinset{r}{-.\textwidth}{t}{-.\textwidth}
   {
   \setlength{\fboxsep}{0pt}%
 \setlength{\fboxrule}{1pt}%
   \fbox{\includegraphics[width=0.4\textwidth]{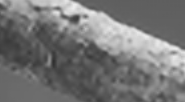}}}
    {\stackinset{l}{-2.5pt}{t}{0pt}
   {
   \setlength{\fboxsep}{0pt}%
 \setlength{\fboxrule}{1pt}%
   \fbox{\includegraphics[width=0.35\textwidth]{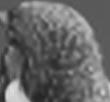}}}
   {\includegraphics[width=\textwidth]{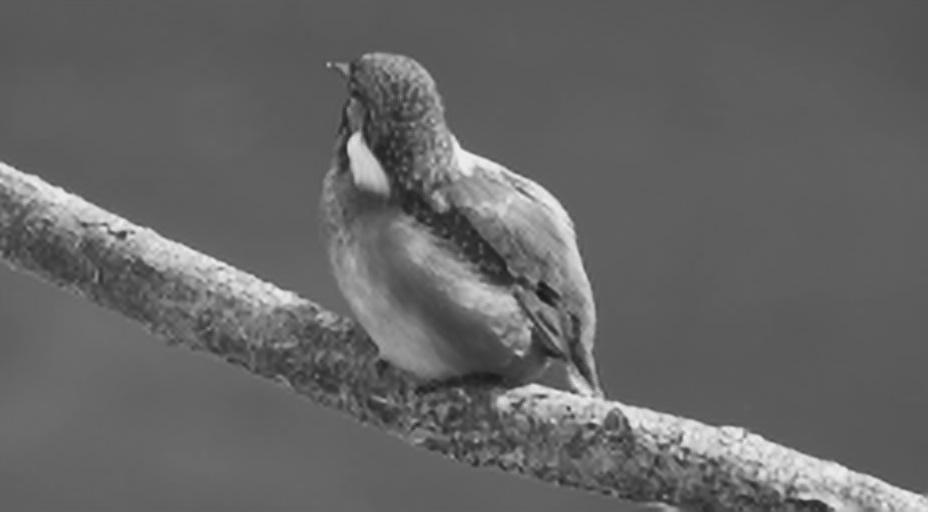}}}
    \caption{RLSP-HL}
    \label{fig:birdRLSPHL}
\end{subfigure}
%
%
%
\hfill 
\begin{subfigure}{0.475\columnwidth}
 \stackinset{r}{-.\textwidth}{t}{-.\textwidth}
   {
   \setlength{\fboxsep}{0pt}%
 \setlength{\fboxrule}{1pt}%
   \fbox{\includegraphics[width=0.4\textwidth]{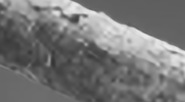}}}
    {\stackinset{l}{-2.5pt}{t}{0pt}
   {
   \setlength{\fboxsep}{0pt}%
 \setlength{\fboxrule}{1pt}%
   \fbox{\includegraphics[width=0.35\textwidth]{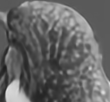}}}
   {\includegraphics[width=\textwidth]{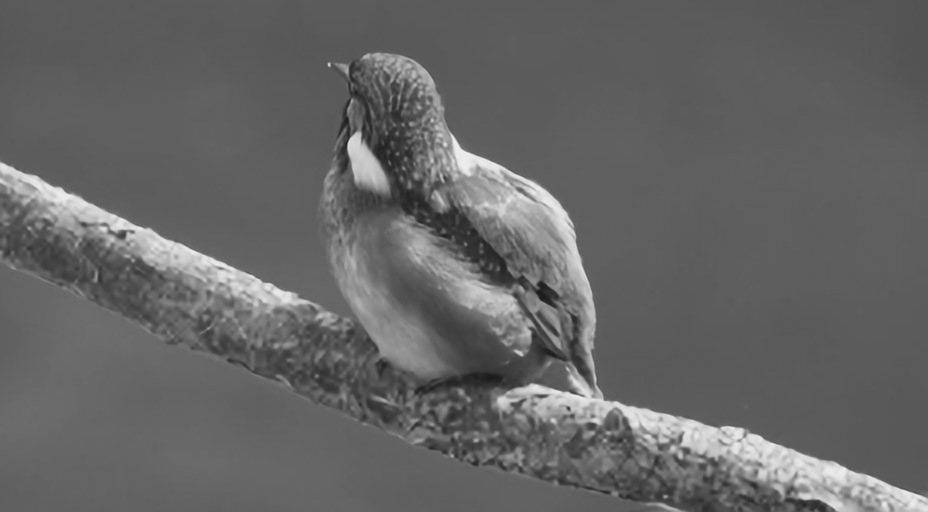}}}
    \caption{RFS3}
    \label{fig:birdRFS3}
\end{subfigure}
\begin{subfigure}{0.475\columnwidth}
 \stackinset{r}{-.\textwidth}{t}{-.\textwidth}
   {
   \setlength{\fboxsep}{0pt}%
 \setlength{\fboxrule}{1pt}%
   \fbox{\includegraphics[width=0.4\textwidth]{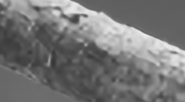}}}
    {\stackinset{l}{-2.5pt}{t}{0pt}
   {
   \setlength{\fboxsep}{0pt}%
 \setlength{\fboxrule}{1pt}%
   \fbox{\includegraphics[width=0.35\textwidth]{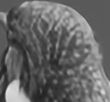}}}
   {\includegraphics[width=\textwidth]{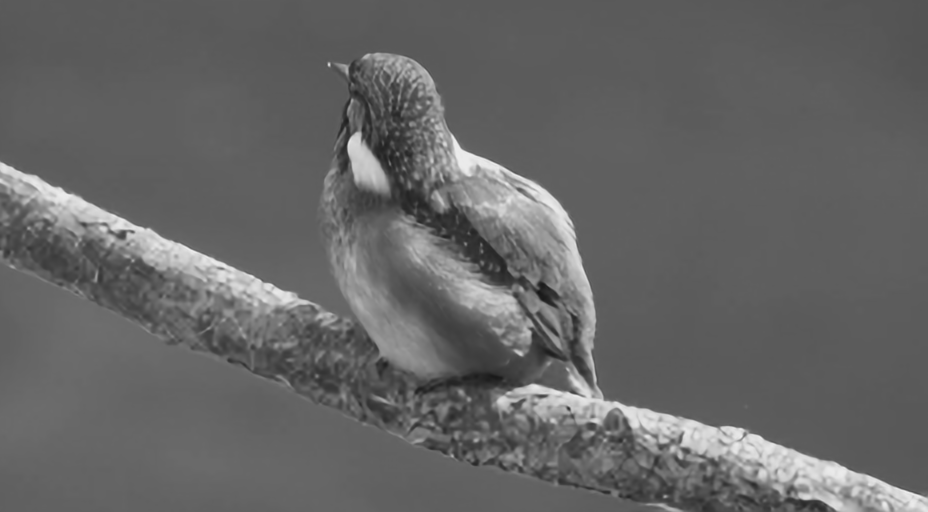}}}
    \caption{RFS7}
    \label{fig:birdRFS7}
\end{subfigure}
\hfill 
\begin{subfigure}{0.475\columnwidth}
 \stackinset{r}{-.\textwidth}{t}{-.\textwidth}
   {
   \setlength{\fboxsep}{0pt}%
 \setlength{\fboxrule}{1pt}%
   \fbox{\includegraphics[width=0.4\textwidth]{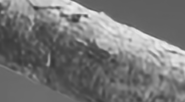}}}
    {\stackinset{l}{-2.5pt}{t}{0pt}
   {
   \setlength{\fboxsep}{0pt}%
 \setlength{\fboxrule}{1pt}%
   \fbox{\includegraphics[width=0.35\textwidth]{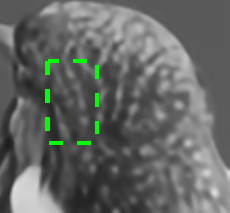}}}
   {\includegraphics[width=\textwidth]{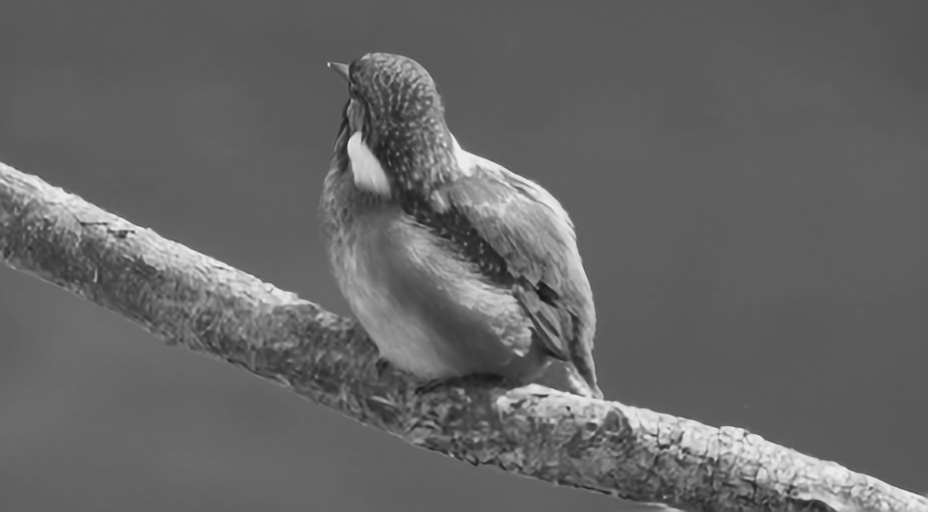}}}
    \caption{MRVSR}
    \label{fig:MRVSRbird}
\end{subfigure}
%
\caption{The 376th frame of the first sequence of Quasi-Static Video Set, reconstructed from methods that are stable by design (non recurrent or under HL). MRVSR presents the best reconstruction.} 
\label{fig:birdstable} 
\end{figure}

\begin{figure}[!htbp]

\centering
   \includegraphics[width=1\linewidth]{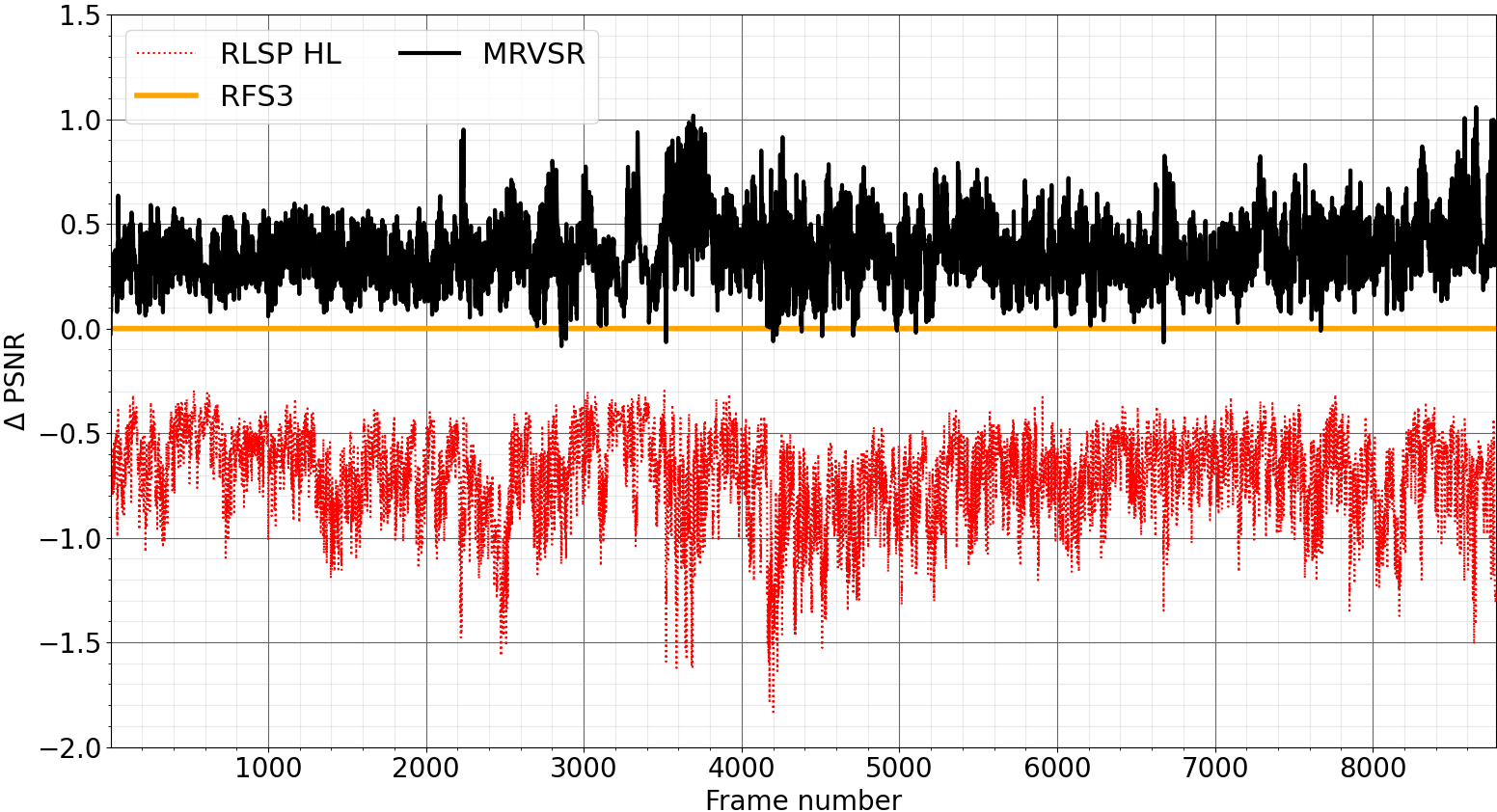}
\caption{Evolution of PSNR on Y channel per frame on \textit{Sequence 1-XL}. We substract the curve of the RFS3 baseline and the graph shows these differences.
}
\label{fig:seq1}
\end{figure}

\par \textbf{SL}: RLSP-SL faces the same issues as existing recurrent networks. After being better than the baseline RFS3 at the beginning of the sequences, it diverges (\cref{fig:recurrentnets}). It generates high frequency artifacts (\cref{fig:birdRLSPSRNL201}) and its performance at the end of the sequences is poor, as shown in \cref{tab:psnrssim} ($-2.09$dB in mean PSNR and $-0.0284$ in mean SSIM compared to RFS3 on the last 50 reconstructions). This proves that SL 
is not enough to prevent the divergence.

\par \textbf{HL}: RLSP-HL also obtains an overall poor performance ($-1.13$dB in average PSNR and $-0.0278$ in average SSIM compared to RFS3 based on all reconstructed frames, according to \cref{tab:psnrssim}). Its reconstruction performance is generally stable on a long sequence (\cref{fig:recurrentnets,fig:seq1}), but the reconstructed image is blurred (\cref{fig:birdRLSPHL}). This is because RLSP-HL is globally constrained to be 1-Lipschitz. Thus, as stated in \cref{USRVSR}, it is poorly suited to the deconvolution task.

\subsection{Performance of the proposed network}

At the beginning of the quasi static sequences (\cref{fig:recurrentnets,tab:psnrssim}) MRVSR can not match RLSP and RSDN, but performs better than the baseline RFS3 and FRVSR. This performance is compatible with the results on Vid4 (\cref{tab:psnrssimVid4}), where MRVSR is 0.56dB behind the unconstrained similar network RLSP. This is due to the Lipschitz constraint on MRVSR, built to ensure its long-term stability at the price of a lower short-term performance.


%
%
%
%
%

\begin{table}[h]

        \centering
        \setlength\tabcolsep{1.75pt}
        \begin{tabular}{| l | l | l |  l |  l |}
        \hline
        RFS3 & FRVSR & RSDN & RLSP & MRVSR \\
        \hline
         26.43 & 26.69 & 27.92 & 27.46 & 26.90 \\
        \hline           
       \end{tabular}
       \caption{Mean PSNR on Y channel of Vid4 dataset. Values for FRVSR, RLSP and RSDN are taken from their publications.}
       
       \label{tab:psnrssimVid4}

\end{table}

\begin{figure}[!htbp]

\centering

    \includegraphics[width=\columnwidth]{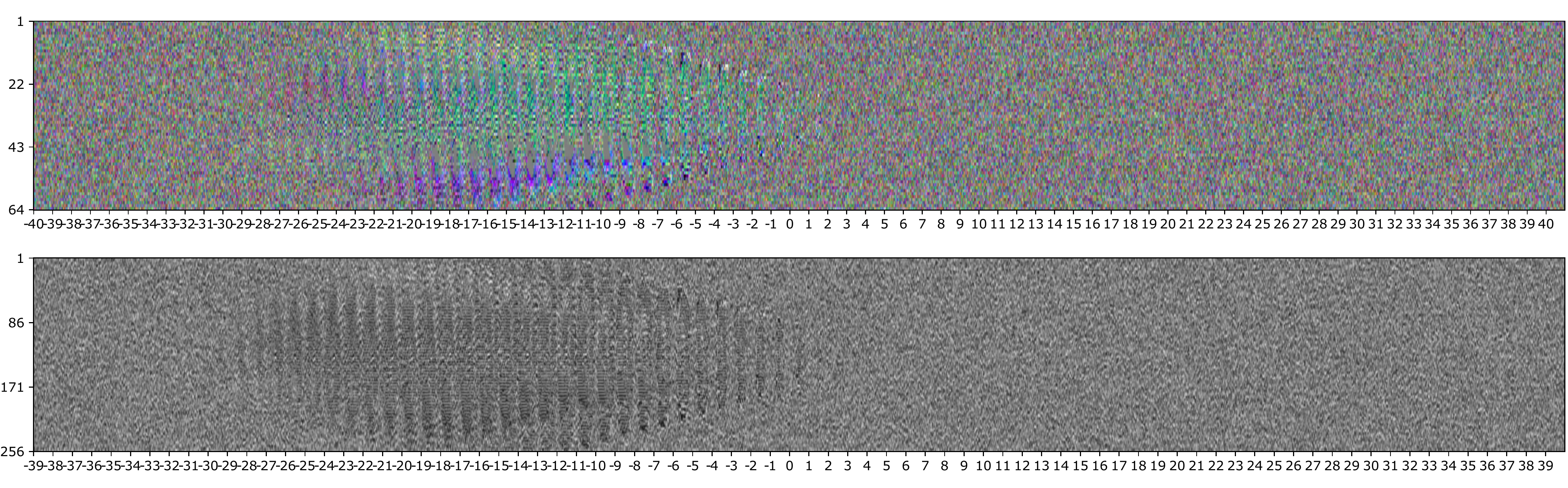}

\caption{Spatio-temporal receptive fields of MRVSR. Top and bottom are respectively input and output (Y channel) sequences. Figures are stretched using nearest neighbor interpolation.}
\label{fig:strfmrvsr}
\end{figure}

\begin{figure}[!htbp]
\centering
\begin{subfigure}{0.475\columnwidth}
    \includegraphics[width=\textwidth]{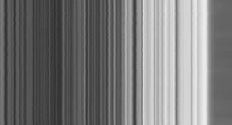}
    \caption{GT}
\end{subfigure}
\hfill
\begin{subfigure}{0.475\columnwidth}
    \includegraphics[width=\textwidth]{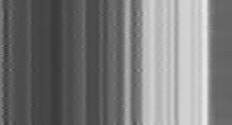}
    \caption{RFS3} 
\end{subfigure} 

\begin{subfigure}{0.475\columnwidth} 
    \includegraphics[width=\textwidth]{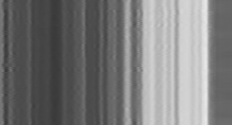}
    \caption{RFS7}
\end{subfigure}  
\hfill 
\begin{subfigure}{0.475\columnwidth} 
    \includegraphics[width=\textwidth]{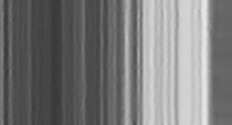}    \caption{MRVSR}
\end{subfigure}

\caption{Temporal profiles from the Y channel of the first sequence of Quasi-Static Video Set.} 
    \label{fig:temporal_profile}
\end{figure}

When considering long-term performance on sequences with low motion, MRVSR gives the best results. \cref{fig:recurrentnets,fig:seq1,fig:MRVSRbird} show that MRVSR does not diverge and does not generate any artifact. According to \cref{tab:psnrssim}, MRVSR achieves the best mean performance on the test set, based on all reconstructed frames as well as focusing on the last 50 reconstructed frames. Because MRVSR and RFS3 take the same number of input frames---namely three---the differences of $+0.58$ dB in average PSNR and +$0.0121$ in average SSIM computed on all reconstructed frames represent the benefit brought by the contractive recurrence map of MRVSR. Moreover, considering that RFS7 takes an input batch of 7 frames, the fact that MRVSR outperforms RFS7 ($+0.39$dB in average PSNR and $+0.0086$ in average SSIM) shows that the temporal receptive field enabled by its contractive recurrence accounts for more than 7 frames. 
This is confirmed in \cref{fig:strfmrvsr}, where the temporal receptive field of MRVSR spans around 28 frames, which is much larger than the usual length (\ie~7) of temporal receptive field of sliding-window based models. Moreover, temporal profiles produced by MRVSR are less noisy and sharper than the ones produced by RFS3 and RFS7. This shows the contractive recurrence map of MRVSR additionally enables increased temporal consistency. Visually speaking, sequences generated by MRVSR present less flickering artifacts than sequences produced by RFS7 and RFS3. \cref{fig:temporal_profile} displays examples of temporal profiles for the first sequence of Quasi-Static Video Set. Finally, MRVSR presents the best long-term reconstruction in terms of visual quality. Some examples can be observed in \cref{fig:birdstable}.

\section{Conclusion}
In this work, we have pointed out the divergence problem of recurrent VSR when facing long sequences with low motion. Existing recurrent VSR networks generate high-frequency artifacts on such sequences. 
To solve this issue, we defined a new framework of recurrent VSR model, based on Lipschitz stability theory. As an implementation of this framework, we proposed a new recurrent VSR network coined MRVSR. We experimentally verified its stability and state-of-the-art performance on long sequences with low motion. As part of our experiments, we introduced a new test dataset of such sequences, namely Quasi-Static Video Set. 

{\small
\bibliographystyle{ieee_fullname}
\bibliography{bib}
}

\end{document}